# Classification of Single Tree Decay Stages from Combined Airborne LiDAR Data and CIR Imagery

Tsz-Chung Wong[1], Abubakar Sani-Mohammed[1], Jinhong Wang[1*], Puzuo Wang[1], Wei Yao[1,2*],

Marco Heurich[3,4,5]

[1] Department of Land Surveying and Geo-Informatics, The Hong Kong Polytechnic University, Hung Hom, Kowloon, Hong Kong

[2] The Hong Kong Polytechnic University Shenzhen Research Institute, Shenzhen, China

[3] Dept. for National Park Monitoring and Animal Management, Bavarian Forest National Park, 94481 Grafenau, Germany

[4] Chair of Wildlife Ecology and Management, Faculty of Environment and Natural Resources, University of Freiburg, Freiburg, Germany

[5] Department of Forestry and Wildlife Management, Campus Evenstad, Inland Norway University of Applied Sciences, Koppang, Norway

- tszchung.wong@connect.polyu.hk, abubakar.sanimohammed@connect.polyu.hk, jinhong.wang@connect.polyu.hk, puzuo.wang@connect.polyu.hk, wei.hn.yao@polyu.edu.hk, marco.heurich@npv-bw.bayern.de

**Abstract:** Understanding forest health is of great importance for the conservation of the integrity of forest ecosystems. The monitoring of forest health is, therefore, indispensable for the long-term conservation of forests and their sustainable management. In this regard, evaluating the amount and quality of dead wood is of utmost interest as they are favorable indicators of biodiversity. Apparently, remote sensing-based machine learning techniques have proven to be more efficient and sustainable with unprecedented accuracy in forest inventory.

*Corresponding author

However, the application of these techniques is still in its infancy with respect to dead wood mapping. This study, for the first time, automatically categorizing individual coniferous trees (Norway spruce) into five decay stages (live, declining, dead, loose bark, and clean) from combined airborne laser scanning (ALS) point clouds and color infrared (CIR) images using three different Machine Learning methods - 3D point cloud-based deep learning (KPConv), Convolutional Neural Network (CNN), and Random Forest (RF). First, CIR colorized point clouds are created by fusing the ALS point clouds and color infrared images. Then, individual tree segmentation is conducted, after which the results are further projected onto four orthogonal planes. Finally, the classification is conducted on the two datasets (3D multispectral point clouds and 2D projected images) based on the three Machine Learning algorithms. All models achieved promising results, reaching overall accuracy (OA) of up to 88.8%, 88.4% and 85.9% for KPConv, CNN and RF, respectively. The experimental results reveal that color information, 3D coordinates, and intensity of point clouds have significant impact on the promising classification performance. The performance of our models, therefore, shows the significance of machine/deep learning for individual tree decay stages classification and landscape-wide assessment of the dead wood amount and quality by using modern airborne remote sensing techniques. The proposed method can contribute as an important and reliable tool for monitoring biodiversity in forest ecosystems.

# 1. Introduction

Forests are precious natural resources that deliver important ecosystem services to the benefit of humankind (Brockerhoff et al., 2017). Therefore, maintaining and monitoring forest health is of great importance for the conservation of the integrity of forest ecosystems. Forests play a critical role in maintaining biodiversity and in the carbon cycle (Paniagua-Ramirez et al., 2021; Houghton 2005; Polewski et al. 2021) and can contribute to reducing the greenhouse effect and consquently alleviating global warming if sustainably managed. However, various diseases and disturbances could cause tree decay, including microorganisms, insect infestation, droughts, wildfires, and extreme weather conditions (Blanchette, 1998; Trumbore et al., 2015; Seibold et al., 2021). For instance, bark beetle attacks have caused large-scale disturbances in recent years (Latifi, 2014; Huang et al., 2020; Hlásny et al. 2021; Sani-Mohammed, et al., 2022). Such areas can harbor large amounts of dead wood, which make the carbon storage assessment very vital. Moreover, dead wood is an important resource for biodiversity (Jonsson et al. 2005; Lassauce et al. 2011; Stokland et al. 2012; Seibold et al. 2015). However, to harbor high biodiversity, the quantity and quality of the dead wood is crucial (Similä et al. 2003; Lassauce et al. 2011; Seibold et al. 2016). Therefore, not just the estimation of the amount of dead wood but also the assessment of the dead wood quality is a decisive precondition for sustainable forest management and conservation of biodiversity. Thus, early detection of declining trees can help identify the origin of bark beetle infestation for quick preventive measures before it spreads widely in the forest. This would significantly improve forest management and protect biodiversity (Abdullah et al., 2019). For standing dead wood, Thomas et al. (1979) developed a framework for classifying different decay stages, which proved to be crucial for the habitat modeling of different groups of species, such as birds, bats, and beetles (Similä et al. 2003; Cousins et al. 2015; Zielewska-Büttner et al. 2018; Kortmann etal. 2018). However, the assessment of the different decay stages still depends on a visual evaluation by field crews.

Conventional forest management methods such as field surveying are very time- and cost-consuming due to the large area of forests. With the advancement of remote sensing technologies, the health of forests can be sensed without on-site measurements in a cost-efficient way with high quality (Lausch et al. 2016). Among various remote sensing techniques, airborne laser scanning (ALS) has been widely used in forestry applications as it can accurately map the 3-dimensional structures of the forests. Based on this, crucial information can be extracted from the point cloud data from the tree to the landscape level (Maltamo et al., 2014; Latifi et al., 2015; Moudrý et al., 2022).

In the past, various studies have been conducted to classify and detect standing dead trees using Light Detection and Ranging (LiDAR) point clouds. Yao et al. (2012a) identified standing dead trees in a temperate forest park using full-waveform lidar data by training a binary support vector machine (SVM) classifier, achieving overall accuracy (OA) of 73%. Polewski et al. (2015) and Polewski et al. (2016) proposed a method of integrating ALS data and aerial infrared imagery to detect standing dead trees by using an active learning approach, reaching an OA of 89%. Wing et al. (2015) analyzed the spatial distribution of snags and provided a better understanding of wildlife snag use dynamics using neighbourhood attribute filtered ALS data followed by individual tree detection. The overall detection rate for snags with DBH >=25 cm was 56% with low commission error rates, while detection rates ranged from 43 to 100%, increasing with the size of the snag. A multitude of research has been conducted for classifying tree species together with standing dead trees using lidar data and multispectral images (Yao, et al, 2012b; Polewski et al.,2020). Kamińska et al. (2018) conducted a species-related individual dead tree detection by combining multi-temporal ALS point cloud and color infrared imagery using a Random Forest (RF) classifier, resulting in an OA of 94.3%. Krzystek et al. (2020) studied large-scale mapping of individual conifers, broadleaf trees, and standing dead trees using lidar data and multispectral imagery. They reached an overall accuracy of better than 90% after training RF and logistic regression models. Marchi et al. (2018) reviewed studies based on the application of airborne and terrestrial laser scanning for the identification of large deadwood components (e.g., snags, logs, stumps). They concluded that efforts should be made to transfer the available single-tree methodologies to an operational level but did not consider further characterizing tree conditions. To the best of our knowledge, no research has been published that quantifies different decay classes of standing dead trees using remote sensing techniques until now. Until recently studies have extracted handcrafted features from point clouds and multispectral images for the classification tasks. Deep learning (DL) methods are currently becoming more and more used as classification tools for similar tasks in recent years. For example, Hamraz et al. (2019) constructed a deep convolutional neural network (DCNN), for binary classification of coniferous and deciduous trees using 2D representations of individual trees and achieved an OA of around 90%. Briechle et al. (2020) also studied the possibility of using the 3D deep neural network (DNN) PointNet++ for classifying tree species and standing dead trees. Their results showed that the 3D PointNet++ approach (OA = 90.2%) outperformed the RF classifier and handcrafted features approach (OA=85.3%). Furthermore, Seidel et al. (2021) compared the performance of the image-based CNN approach and point cloud-based PointNet approach for tree species classification tasks. Their study revealed that the CNN approach achieved higher accuracy and computational efficiency than the PointNet approach. However, these studies have only focused on utilizing LiDAR data and multispectral images for classifying tree species or standing dead trees. Here we propose that these kinds of data could be applied to broader applications, such as categorizing individual trees into several decay stages rather than just living and dead trees.

However, to the best of our knowledge, the possibility of using ALS data and multispectral aerial images for classifying tree decay stages is yet to be studied, especially using Machine Learning (ML). Therefore, the objectives of this study are: (1) to define five tree decay stages that could be detected by ALS and multispectral images, (2) to generate two datasets (3D multispectral colorized point clouds and 2D side-view projected images) of individual tree decay stages as a basis for their classification (training), by combining ALS data and CIR aerial imagery, and (3) to assess the performance of three ML/DL algorithms (KPConv, CNN, and RF) in classifying the five decay stages of individual trees.

## 2. Materials

### 2.1 Study area

The experiments were carried out in the Bavarian Forest National Park (BFNP) (49° 3' 19" N, 13° 12' 9" E), located in Germany and along the border with the Czech Republic (Figure 1). Established as the first National Park in Germany, the park covers an area of approximately 24,000 ha with altitudes ranging from 650 to 1453 m above sea level (Heurich et al., 2010). The park is dominated by Norway Spruce (*Picea abies*), European Beech (*Fagus sylvatica*), European Silver fir (*Abies alba*), and other Spruce trees (van der Knaap et al., 2020). Despite experiencing severe thunderstorms in the 1980s and a bark beetle attack during the 1990s (Lausch et al., 2013), the park was protected from human intervention due to the policy of the authority, resulting in a total tree mortality rate of 22% (Müller & Job, 2009).

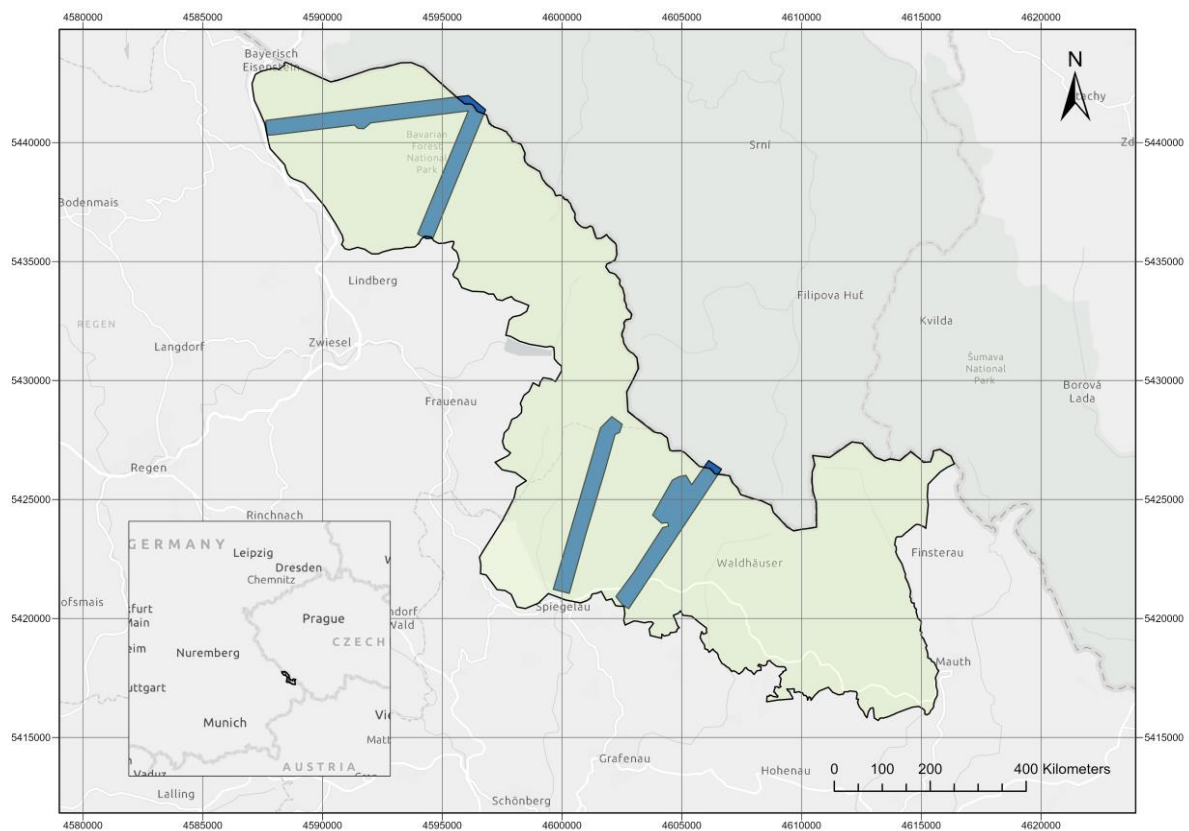

**Figure 1.** Map of the Bavarian Forest National Park (BFNP) and the three transects of ALS point clouds (marked in blue)

**2.2 Data acquisition**

ALS point clouds and color infrared images were acquired in 2016 for the BFNP administration within the framework of the data pool initiative for the Bohemian Forest Ecosystem (Latifi et al., 2021). Detailed information on the data is described in sections 2.2.1 and 2.2.2 respectively.

**2.2.1 ALS data**

Three transects of ALS point clouds (leaf-on) were acquired on 18$^{th}$ August 2016 with an LMS-Q680i - 400 kHz scanner in a flight operated by the Milan Geoservice GmbH, at an altitude of 300 m above sea level. This exercise covered a total area of 13.5 square km with an estimated point density of 70 points per square meter. The raw data was processed and corrected to LAZ 1.2 format (with X, Y, Z, and Intensity), while the WGS84/UTM coordinates were calculated with a transformation fitted according to the Gauss-Krüger coordinate system. Figure 2 shows the top view of the three transects of ALS point clouds visualized by intensity values.

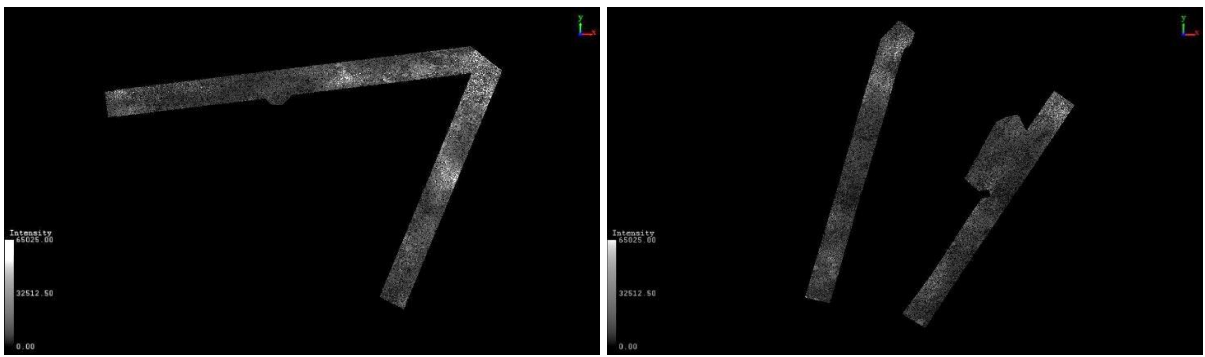

**Figure. 2** The transect of ALS data displayed by the laser intensity (left: transect 1; right: transect 2 and 3)

**2.2.2 Color infrared images**

Three-band CIR aerial images (G, R, NIR) covering the entire BFNP park were acquired by a Leica DMC 122 camera from a Cessna 207 Aircraft. The flight was conducted on 23$^{rd}$ June 2016 by a Berlin-based remote sensing company, ILV Remote Sensing GmbH, at an average altitude of 2918 m above sea level. The longitudinal overlap of the flight was 75%, while the cross-coverage was 60%. The camera was calibrated in 2014, with a focal length of 120 mm, and was used to capture the images with a ground resolution of 20 cm. Similar to the point cloud data, the images were geo-referenced according to the Gauss-Krüger coordinate system. The images were radiometrically corrected and orthorectified. Figure 3 shows an example of a cropped CIR image. Unhealthy trees are identified by the green band of this false-color image because defoliated trees less reflect near-infrared radiation. In contrast, healthy trees appear in red as they reflect a high proportion of near-infrared radiation (Knipling, 1970).

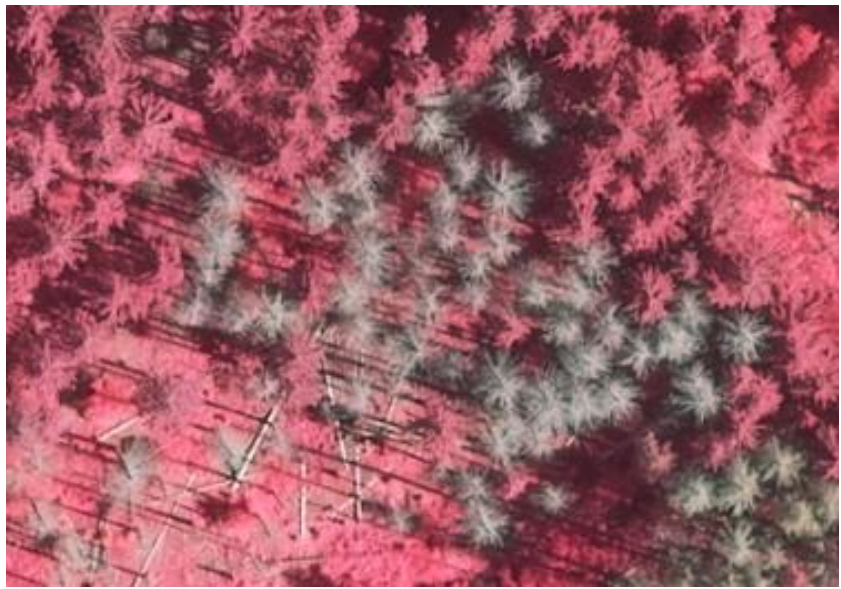

**Figure 3.** A sample CIR aerial image captured in BFNP

**2.3 Definition of tree decay stages**

Thomas et al. (1979) defined nine different decay stages, starting from the "live" to the "stump" stage. However, in this study, we are considering only the first five stages of decay (live, declining (in our case, this stage is equally dead but at an early stage, which can be identified by the change in foliage color and canopy density), dead, loose bark, and clean) proposed by Thomas et al. (1979) for experiments. The tree decay levels we defined are inspired by Thomas et al.'s (1979) description, while they are not entirely the same (Figure 4). For easier differentiation, we renamed these five stages starting from Decay Level 1 to Decay Level 5, respectively. Figure 4 illustrates multispectral point clouds of individual tree at the five decay stages.

| Live | | Dead | | |
|---|---|---|---|---|
| Decay stages | | | | |
| 1. Live | 2. Declining | 3. Dead | 4. Loose bark | 5. Clean |
| 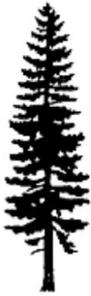 | 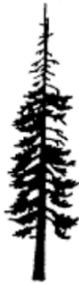 | 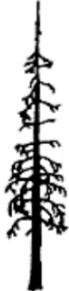 | 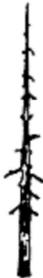 | 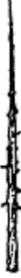 |
| Description reflects our case | | | | |
| No decay; tree is healthy. No deformity; Have thick canopy. Identified by structural thickness and foliage color (red to brown in CIR imagery) | Dead at early stage with very little growth deformity; identified by the change in foliage color (green to gray); Very little or no defoliation; canopy density is similar to that of live tree. | Dead at a late stage; large defoliation seen; more than 50% of twigs available; root firm; tree identified by the change in foliage color and structure. | (Almost) No needles/ twigs; more than 50% of branches lost; loose bark; top usually broken; roots stable | No foliage; almost all branches/ bark absent; often have broken tops; roots of larger trees stable. |
| Visualization by corresponding multispectral ALS point clouds | | | | |

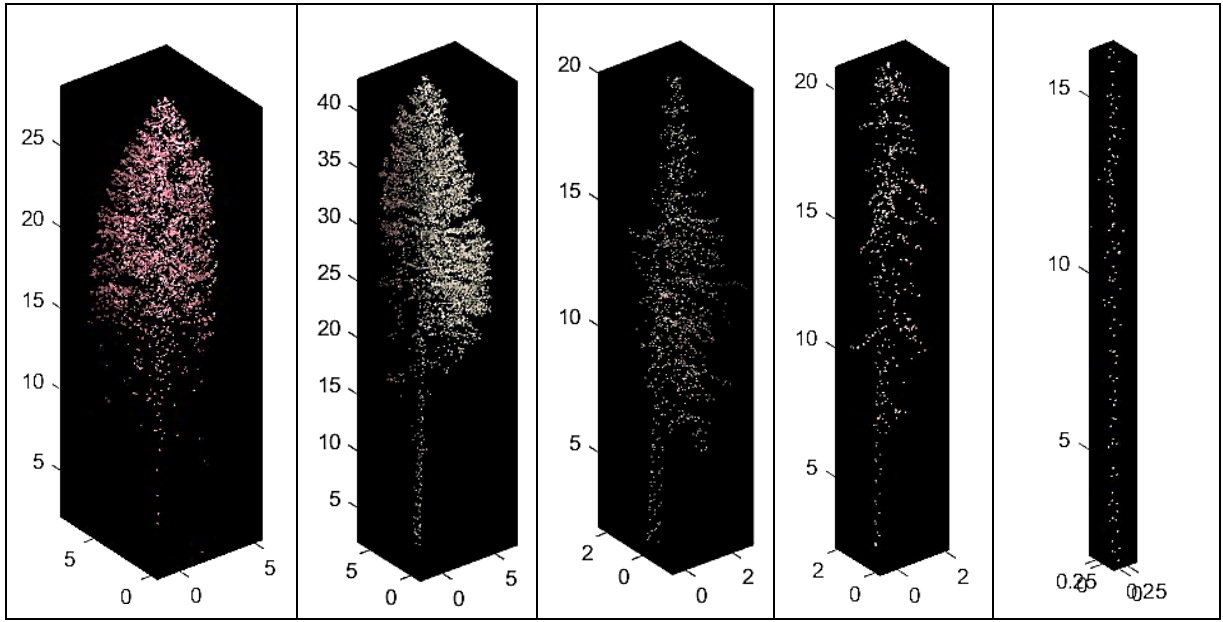

**Figure 4.** Classification standard for decay stages for wildlife (coniferous) standing dead tree (Thomas et al., 1979) and corresponding decay of individual spruce trees from multispectral point clouds.

## 2.4 Reference data curation for tree decay analysis

Reference data were collected by combing manual interpretation of multispectral point clouds of individual trees with expert field experience according to the defined stages of wildlife tree decay described in section 2.3. This important exercise was conducted with expert advice and the additional support of the CIR imagery. Several reference plots in the BFNP were considered in such a way that there was a balance in the diversity of decay levels. The terrain of the forest ranges between ~ 600m and ~1435 m while the height distribution of trees ranges between ~5 m and ~40 m. It is noted that such a manual point cloud labeling method was used due to a lack of complete and precise field data. In total, 1,030 samples were interpreted and labeled. Specifically, the number of labeled samples for the classes was 233, 167, 236, 239, and 155, for decay Level 1, decay Level 2 decay Level 3, decay Level 4, and decay Level 5, respectively.

# 3. Methodology

## 3.1 Overview

After data acquisition, the data was pre-processed and prepared for the classification algorithms. We merged the ALS point clouds with the CIR images and conducted a semi-automatic individual tree segmentation. Since there is a significant difference in the number of data points for individual trees at different levels of decay (Tabel 1), for 3D-point-based classification, we chose KPConv, a deep learning model that does not require a fixed number of input points for training. Also, we transformed each individual tree point cloud into four 2D projected images via multi-view orthographic projection to form an image dataset for CNN and RF classification. These 2D projections were the dataset used for performing image-based classification using CNN and RF approach. These two sets of data were then trained in the three ML/DL algorithms. Finally, the trained models were tested, and the results were evaluated to assess their accuracy. Figure 5 illustrates the general workflow of the methodology.

**Table 1.** Minimum point count, maximum point count, and mean point count of each class

| Class | minPointCount | maxPointCount | meanPointCount |
|---|---|---|---|
| Level 1 | 845 | 10295 | 3346 |
| Level 2 | 1586 | 14405 | 4785 |
| Level 3 | 644 | 5963 | 2139 |
| Level 4 | 121 | 2799 | 637 |
| Level 5 | 16 | 1418 | 161 |

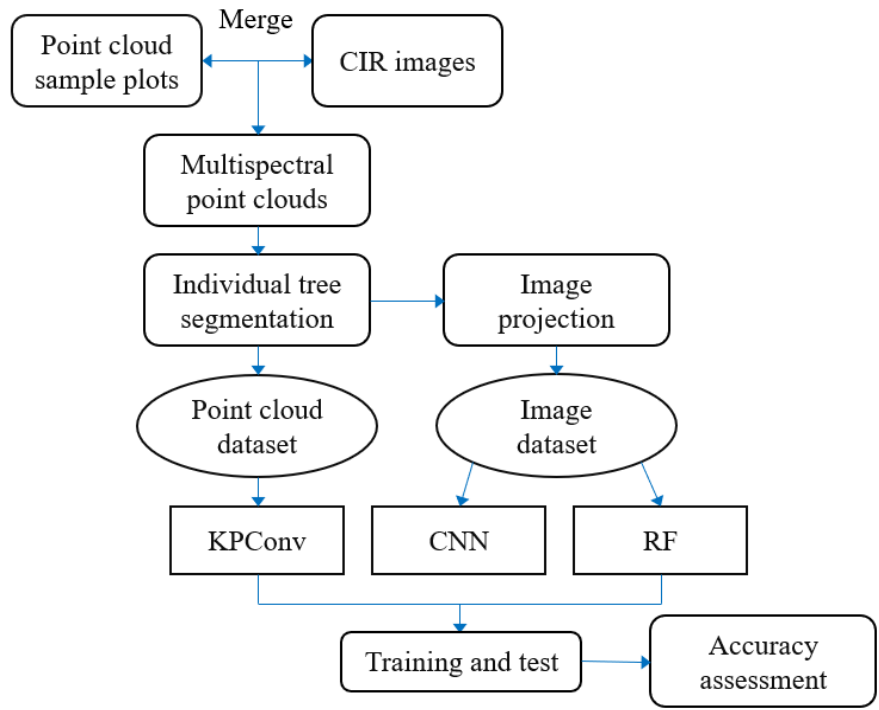

**Figure 5.** General workflow of the methodology

## 3.2 Multispectral point cloud curation

We generated colors for the point cloud of the selected land parcel with a radius of 30m by combining ALS point cloud (X, Y, Z, I) with its corresponding georeferenced CIR image (NIR, R, G). Thus, the ALS point cloud was colorized by the georeferenced CIR images (orthophotos) such that LiDAR points at various spatial positions are assigned with the exact corresponding colors (pixel values) of the image pixels where the LiDAR points fall into. For instance, the NIR, R, and G values of the CIR imagery at ($x_i$, $y_i$) would be applied to all the ALS point clouds at ($X_i$, $Y_i$). This combination generated three more values to the existing four resulting in seven values (X, Y, Z, I, NIR, R, G). (Figure 6).

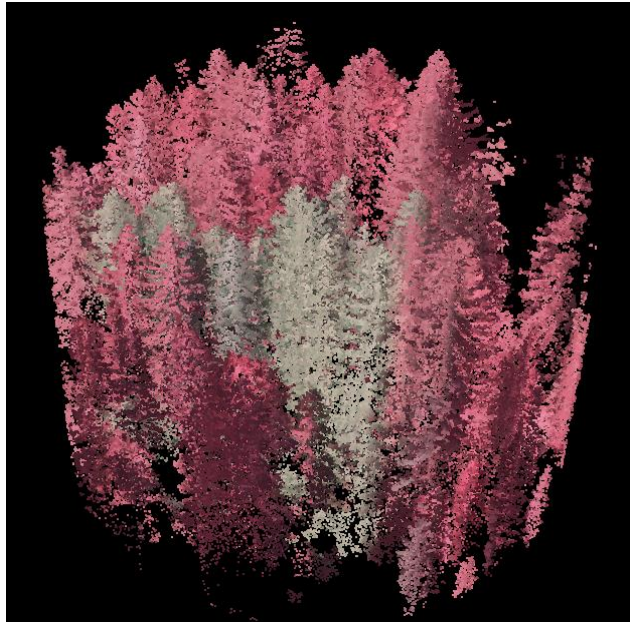

**Figure 6.** A sample plot of combined ALS data and CIR image resulting in multispectral colorized point clouds.

**3.3 Individual tree segmentation**

A semi-automatic individual tree segmentation approach was conducted to select appropriate samples for training. This was after color values have been added to the ALS point clouds as explained in section 3.2.1. In this section, we conducted an automatic individual tree segmentation following the algorithm proposed by Li et al. (2012). The ground points were classified in this stage using an improved progressive TIN densification filtering algorithm (Zhao et al., 2016). We then normalized the point clouds with reference to the ground points to remove the effects of topographic relief. One critical parameter of this algorithm is the 2D Euclidean distance threshold between two adjacent trees because inappropriate thresholds could cause under-segmentation or over-segmentation. The threshold value was set as 0.5 m due to the narrow tree crowns, which led to a shorter distance between two neighboring trees. Nevertheless, the segmented trees from the automatic approach still needed some final evaluation.

Thus, the automatically segmented trees were manually edited and refined in an expert interactive mode by deleting extra point clouds that do not belong to a single tree. Besides, the unwanted point clouds of bushes and young trees around selected individual trees were manually cleaned (Figure 7). And trees with incomplete point clouds were not selected. Overall, a total of 1030 individual tree samples were extracted and constituted the data set. Figure 8 illustrates a sample plot of individual tree segmentation result.

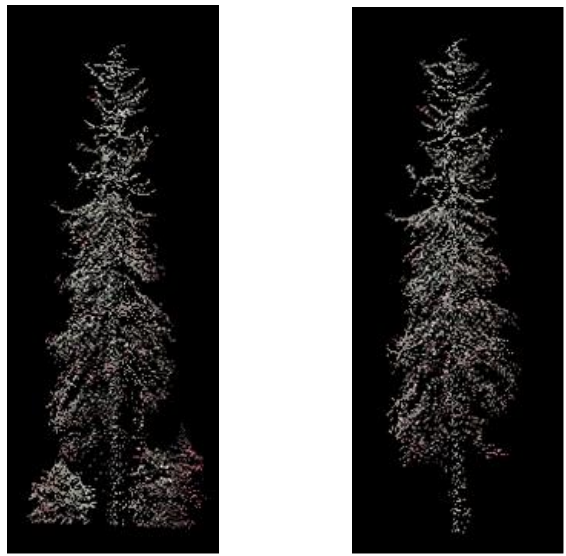

**Figure 7.** A sample of automatic individual tree segmentation (left) and semi-automatic individual tree segmentation (right)

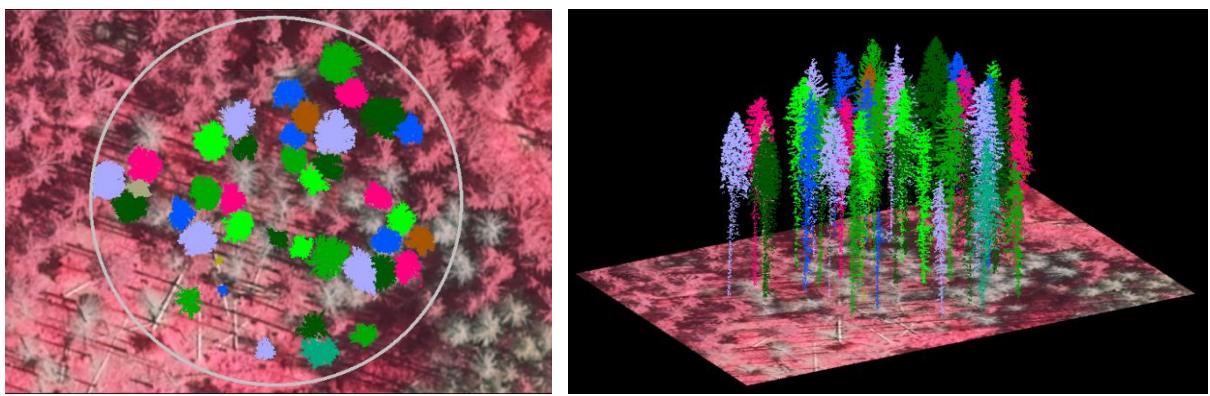

**Figure 8.** A sample plot of individual tree segmentation result

**3.4 Image projection**

In addition to the 3D point clouds of individual trees, an image dataset was generated from the 3D point clouds for classification. Each individual tree point cloud was transformed into four 2D projected images via multi-view orthographic (parallel) projection; the four mutually perpendicular side-views that are produced by four mutually perpendicular planes of projection (Front view 0°, Left-side view 90°, Rear view 180°, and Right-side view 270°) as shown in Figure 9; and this was consistent for each tree. This operation, while it serves to increase the number of training datasets, could equally minimize information loss caused by 3D to 2D representation (Seidel et al., 2021). Table 2 summarizes the distribution of samples in both the 3D point cloud dataset and the 2D image dataset.

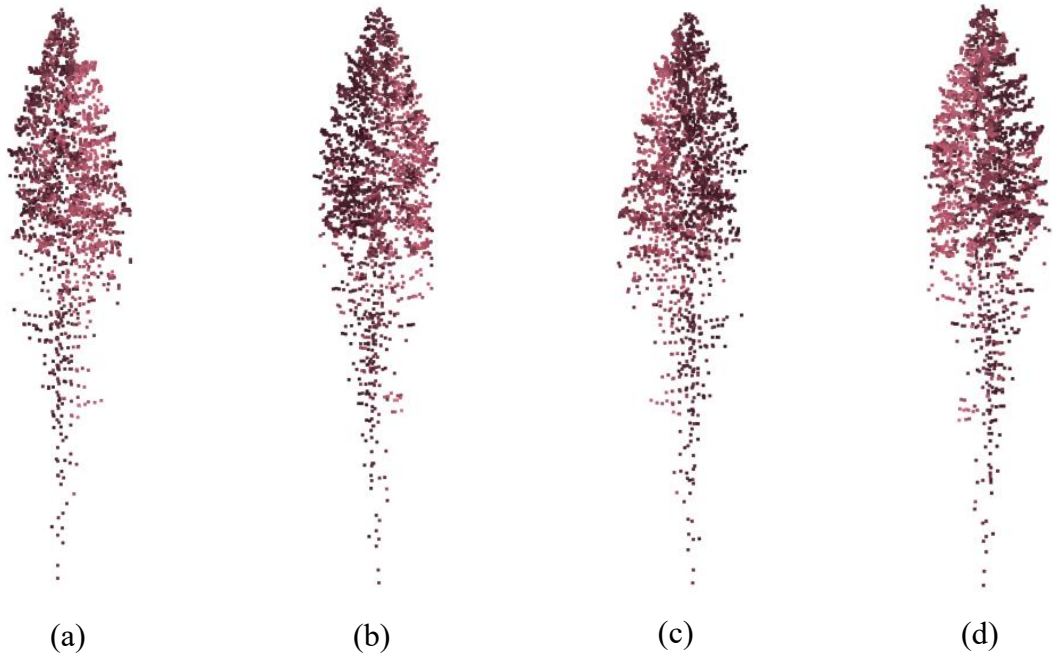

(a)  (b)  (c)  (d)

**Figure 9.** CIR-based images of the same live tree projected from (a) front view, (b) left-side view, (c) rear view, and (d) right-side view

**Table 2.** Number of tree samples in point cloud dataset and image dataset

| Class label | Point cloud dataset | Image dataset |
|---|---|---|
| Level 1 (live) | 233 | 932 |
| Level 2 (declining) | 167 | 668 |
| Level 3 (dead) | 236 | 944 |
| Level 4 (loose bark) | 239 | 956 |
| Level 5 (clean) | 155 | 620 |
| **Total** | **1030 samples** | **4120 images** |

## 3.5 Model architecture

### 3.5.1 KPConv

KPConv (Thomas et al., 2019) introduces a new point convolution operator called Kernel Point Convolution, which uses a set of kernel points to define a local neighborhood around each point in the point cloud. The convolution weights are then applied to the points within this local neighborhood. The number and location of the kernel points can be learned by the network, which allows KPConv to be flexible and deformable. Figure 10 illustrates a 2D Kernel Point Convolution.

The input point cloud $\{Pi \mid i = 1, \ldots, n\}$ is represented by its $x$, $y$, and $z$ coordinates with additional feature channels such as color and intensity. After several layers of convolution, KPConv produces the probability result of predefined categories.

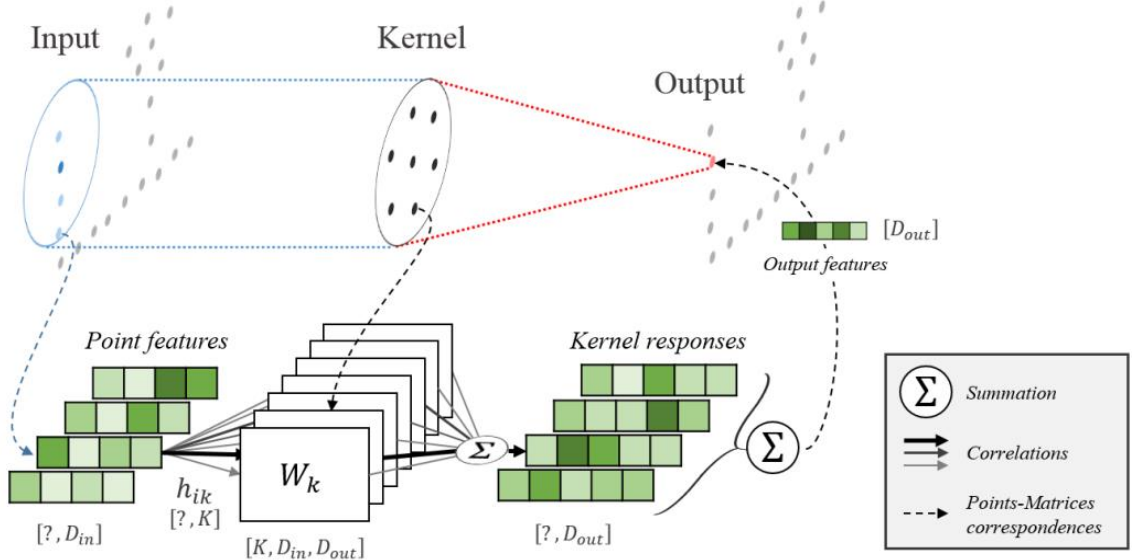

**Figure 10.** An illustration of KPConv on 2D points (Thomas et al., 2019).

**Data preparation**

The minimum, maximum, and mean point count of each class are listed in Table 1. Due to the sparsity of point data in each sample, subsampling would affect its local structure. Therefore, we have removed the subsampling step. Moreover, the point's feature - intensity, NIR, R, and G are normalized into the same range.

**3.5.2 CNN**

Unlike KPConv, CNN takes 2-dimensional images as input. In general, CNN consists of an input layer, multiple hidden layers, and an output layer (Figure 11). The input layer takes images of the same image size, generating a shape of (number of inputs) × height × width × (number of channels). The hidden layers include several convolution layers, pooling layers, a flatten layer and fully connected or dense layers. The convolution layers extract features from the images with the aid of kernels, abstracting the images to a feature map with the shape of (number of inputs) × (height of feature map) × (height of feature map) × (number of channels). The pooling layer follows a convolution layer to reduce the dimensions of the feature maps.

The features pooled at each new location are obtained by max pooling. The most widely used pooling method for image classification is max pooling. Max pooling was applied using Equation (1):

$$f_m(v) = max_i v_i \qquad (1)$$

where the vector $v$ is a single $P$-dimensional column in a $P \times k$ matrix reduced by a pooling operation $f$ (Boureau, 2010). The feature maps extracted in the final convolution and pooling layer are flattened into a vector using the flattening layer before connecting to the fully connected layers to perform classification. We built CNN from scratch for our experiments. Further information on our hyperparameters for the model is explained in section 4.2.

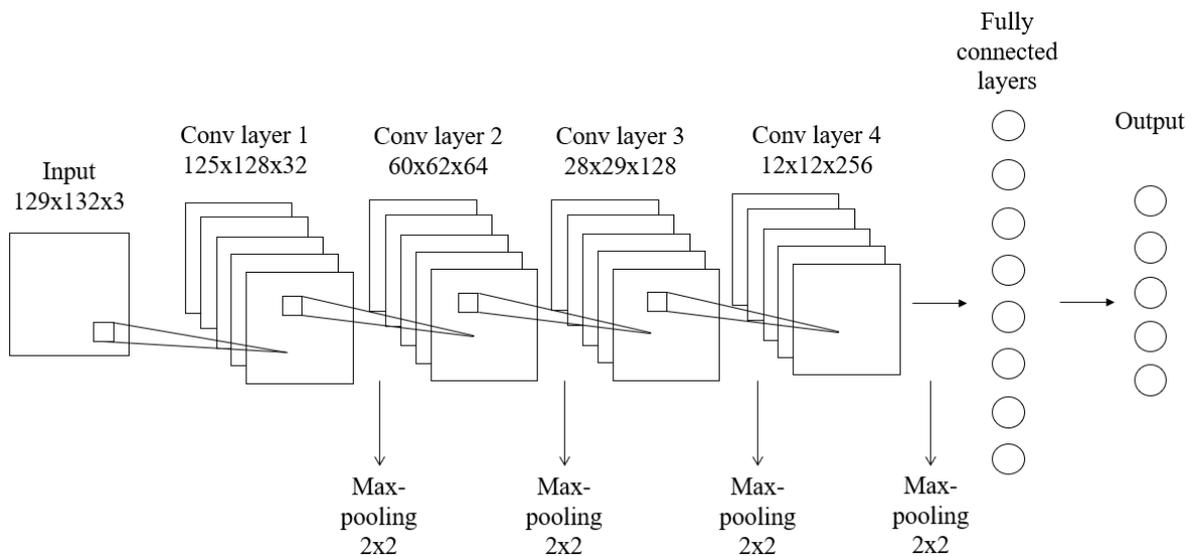

**Figure 11.** The architecture of the CNN model built for this study.

**Dataset preparation**

All images were downscaled by 0.2, generating images with sizes of 129 x 132 pixels with normalized pixel values of the three channels.

**3.5.3 Random Forest**

Random Forest (RF) is a supervised ensemble learning algorithm proposed by Breiman

(2001). Ensemble learning aims at improving model performance by running a base learning algorithm several times. Combining the hypotheses generated by the algorithm at each time, a voted hypothesis could be obtained, which usually leads to a better prediction (Dietterich, 2002). RF is an improved version of the decision trees algorithm by reducing the variance of the model. It generates multiple decision trees randomly, then combines and averages the outputs of the decision trees to obtain a final prediction. Figure 12 illustrates the model architecture of a random forest classifier.

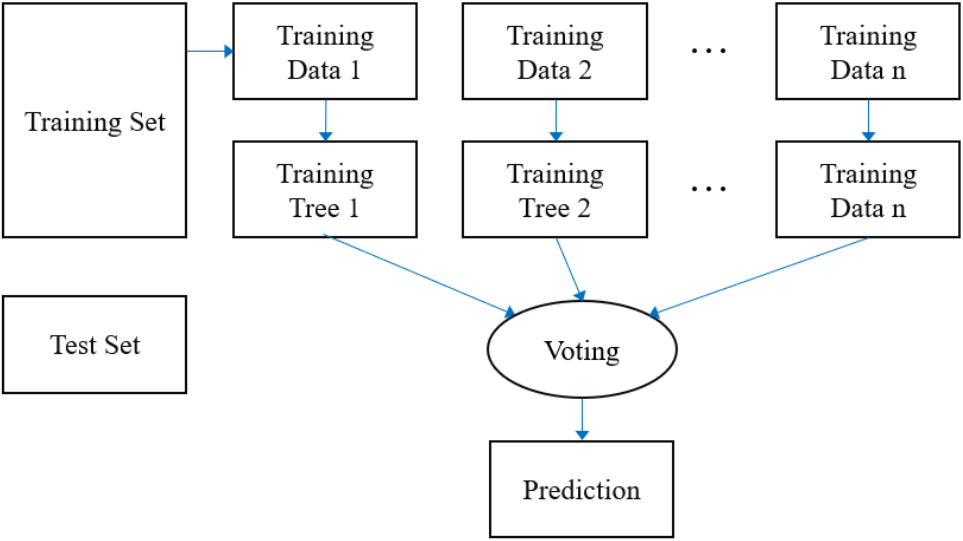

**Figure 12**. The model architecture of an RF classifier

**Dataset preparation**

The processing on the image dataset for RF classification was the same as that for CNN image classification, which is mentioned in section 3.5.2.

**Feature extraction**

Unlike DL, RF being a traditional ML algorithm requires handcrafted feature. We tested typical handcrafted features including Hu moments (Hu, 1962), Haralick texture (Haralick et al., 1973), Histogram of Hue Saturation Value (HSV), Histogram of Oriented Gradients (HOG) (Dalal & Triggs, 2005), and Harris Corner Detection (Harris & Stephens, 1988). Figure 13 shows the ranking of the feature importance.

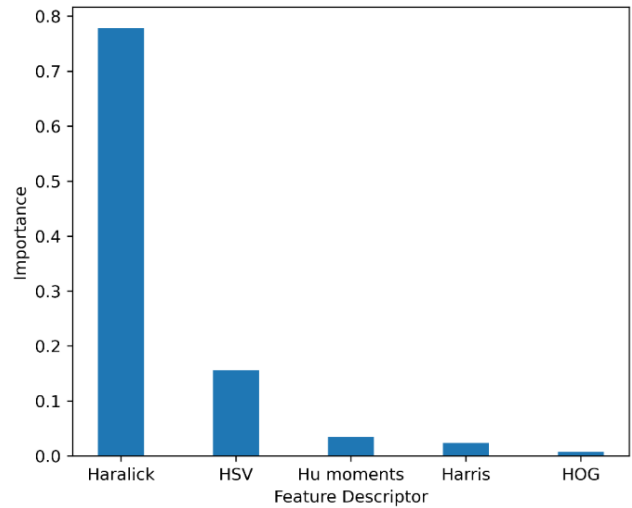

**Figure 13.** Feature importance ranking for RF classification.

The top three feature sets (Haralick texture, HSV histogram, and Hu moments) are selected as RF classification input features and extracted from each processed image. These features were combined into a global feature vector as the input for the RF Classifier. Figure 14 depicts a 2D feature map (including 1$^{st}$ and 2$^{nd}$ principal components) generated by Principal Component Analysis (PCA) with the five classes of tree decay.

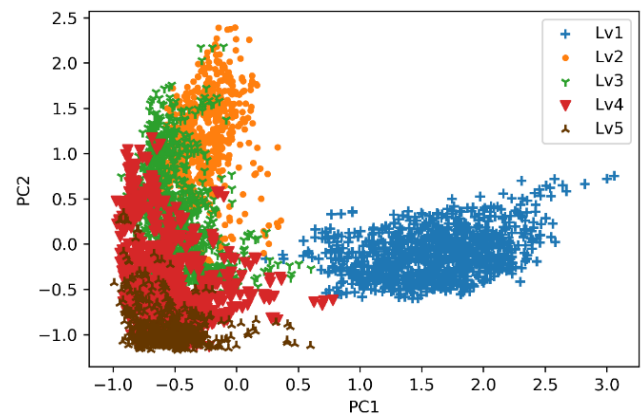

**Figure 14.** A 2-D feature space generated after feature extraction and PCA.

## 3.6 Model evaluation

The performance of our models' classification was evaluated based on the overall accuracy (OA), F1- scores, and Cohen's kappa coefficient (κ). The OA is expressed as the sum of correctly classified features relative to the total number of features of the confusion table as represented in Equation (2):

$$OA = \frac{TP+TN}{TP+TN+FP+FN} \tag{2}$$

where TP, TN, FP, and FN represent true positive, true negative, false positive, and false negative respectively. F1-score is the weighted average of recall and precision which can be computed as expressed in Equation (3):

$$F1 = 2 \cdot \frac{recall \cdot precision}{recall+precision} \tag{3}$$

where $Recall = \frac{TP}{TP+FN}$ and $Precision = \frac{TP}{TP+FP}$. Cohen's kappa coefficient ($\kappa$) measures the agreement between two classifiers by considering the probability of the observed agreement and the probability of agreement by chance (Vieira, 2010). This is expressed in Equation (4):

$$\kappa = \frac{p_o - p_c}{1 - p_c} \tag{4}$$

where the Probability of Observed Agreement is calculated by $p_o = \frac{\sum x_{ii}}{N}$ and the Probability of Agreement by Chance is derived from the formula $p_c = \frac{\sum x_{i+} x_{+i}}{N^2}$.

## 4. Experiments and results

### 4.1 Experimental design

KPConv process point clouds with user-defined feature channels. Thus, we designed four scenarios for our experiments. The utilized features are (X, Y, Z), (X, Y, Z, NIR, R, G), (X, Y, Z, I), and (X, Y, Z, I, NIR, R, G), respectively. We applied these dynamics to perform a sensitivity analysis relating to the color effect as well as the intensity in our evaluations. The 5-fold cross-validation technique is used for robust classification training process, which is also used for CNN and RF classification. Data augmentation was applied to modify the training dataset slightly to avoid model over-fitting. The techniques used for data augmentation include point cloud random rotation, removal, and jittering with Gaussian noise.

### 4.2 Hyperparameter settings

#### i) KPConv

Table 3 shows the hyperparameter - settings considered when training the KPConv DL model. The hyperparameters were adjusted by a manual search approach to achieve improved classification performance.

**Table 3.** KPConv hyperparameters and corresponding values

| Hyperparameter | Value | Description |
|---|---|---|
| inputChannelSize | 7 | Number of input channels per sample (X, Y, Z, I, NIR, R, G) |
| numEpochs | 100 | Number of epochs |
| learnRate | 0.005 | Initial learning rate |
| batchLimit | 50000 | Max number of points per batch |

| numKernelPoints | 15 | Number of kernel points |

### ii) CNN

Generally, our network consists of four convolutional layers, four max-pooling layers, one flattened layer, and two dense layers. We built this CNN model architecture from scratch as illustrated in Table 4. The input shape of each CIR image was (129, 132, 3). The first convolutional layer consists of 32 filters with a kernel size of 5 x 5. This kernel size was used due to the relatively large input of the processed image dataset. The second convolutional layer consists of 64 filters with a kernel size of 3 x 3, followed by the third convolutional layer, which comprises 128 filters with the same kernel size. Lastly, the fourth convolutional layer was made up of 256 3 x 3 filters. Max pooling layers with 2 x 2 filters were added between every two convolutional layers. The Rectified Linear Unit (ReLU) was used as the activation function throughout the network. To complete the convolutional architecture, a flattened layer was used to flatten the three-dimensional feature to a one-dimensional feature for easy pass to the dense layers. The two dense layers have the shape of 256 and 5, respectively. The ReLU was used for the first dense layer, while the softmax activation function was used for the second dense layer (output layer). The softmax activation function helps transform the input vector into a probability vector for performing classification. The output shape (none, 5) in the output layer represents the five classes of tree decay levels.

**Table 4.** Description of the CNN model architecture

| Layer (type) | Output Shape | Param # |
|---|---|---|
| conv2d (Conv2D) | (None, 125, 128, 32) | 2432 |
| max_pooling2d (MaxPooling2D) | (None, 62, 64, 32) | 0 |
| conv2d_1 (Conv2D) | (None, 60, 62, 64) | 18496 |

| Layer | Output Shape | Param # |
|---|---|---|
| max_pooling2d_1 (MaxPooling2D) | (None, 30, 31, 64) | 0 |
| conv2d_2 (Conv2D) | (None, 28, 29, 128) | 73856 |
| max_pooling2d_2 (MaxPooling2D) | (None, 14, 14, 128) | 0 |
| conv2d_3 (Conv2D) | (None, 12, 12, 256) | 295168 |
| max_pooling2d_3 (MaxPooling2D) | (None, 6, 6, 256) | 0 |
| flatten (Flatten) | (None, 9216) | 0 |
| dense (Dense) | (None, 256) | 2359552 |
| dense_1 (Dense) | (None, 5) | 1285 |

**Total params: 2,750,789**
**Trainable params: 2,750,789**
**Non-trainable params: 0**

We used the sparse categorical cross-entropy loss function. The number of training epochs per iteration was set to 20 and the loss was optimized with Adam optimizer.

**iii) Random Forest**

For the RF classifier, we applied automatic grid search for tuning the hyperparameters. We selected a combination with the highest cross-validation score as the final hyperparameters illustrated in Table 5.

Table 5. Random Forest hyperparameters and corresponding values

| Hyperparameter | Value | Description |
|---|---|---|
| n_estimators | 800 | Number of estimators |
| max_depth | 64 | Maximum depth of the decision tree |
| random_state | 42 | Parameter to control the random number generator |
| class_weight | balanced | Parameter to adjust classification weights |

## 4.3 Results

### 4.3.1 KPConv

Table 6, 7, 8 and 9 summarize the OAs, Kappa scores, and F1-scores of our three experiments using KPConv. Overall, scenario 4 (X, Y, Z, I, NIR, R, G) achieved the best results with average OA and Kappa scores reaching 88.8% and 0.859 respectively, throughout the 5-fold cross-validation, indicating substantial agreement. In terms of tree decay levels, level 1 outperformed the others with the highest average F1-score reaching 0.994, while level 4 achieved the lowest average F1-score of 0.825. Scenario 1 (X, Y, Z) had the worst results, with average OA and Kappa scores of only 78.9% and 0.735, respectively. Hence, we believe that both intensity and color are beneficial to the classification.

**Table 6.** The OA, Kappa score, and F1-score of KPConv classification (scenario 1)

|  | OA | Kappa score (κ) | F1-score |  |  |  |  |
|---|---|---|---|---|---|---|---|
|  |  |  | Level 1 | Level 2 | Level 3 | Level 4 | Level 5 |
| 1st iteration | 0.804 | 0.754 | 0.759 | 0.684 | **0.8** | **0.843** | **0.954** |
| 2nd iteration | 0.782 | 0.725 | 0.771 | 0.742 | 0.732 | 0.774 | 0.909 |
| 3rd iteration | 0.787 | 0.73 | **0.8** | 0.667 | 0.769 | 0.776 | 0.931 |
| 4th iteration | 0.752 | 0.690 | 0.790 | 0.717 | 0.693 | 0.785 | 0.772 |
| 5th iteration | **0.822** | **0.777** | **0.8** | 0.779 | 0.795 | 0.828 | 0.925 |
| **Average** | 0.789 | 0.735 | 0.784 | 0.718 | 0.758 | 0.801 | 0.898 |

(The best results of each parameter are marked in bold text.)

**Table 7.** The OA, Kappa score, and F1-score of KPConv classification (scenario 2)

|  | OA | κ | F1-score |  |  |  |  |
|---|---|---|---|---|---|---|---|
|  |  |  | Level 1 | Level 2 | Level 3 | Level 4 | Level 5 |
| 1st iteration | **0.885** | **0.855** | 1 | 0.875 | **0.822** | **0.845** | 0.892 |
| 2nd iteration | 0.854 | 0.816 | 0.978 | 0.881 | 0.733 | 0.788 | **0.925** |
| 3rd iteration | 0.874 | 0.84 | 0.989 | 0.862 | 0.804 | 0.826 | 0.912 |
| 4th iteration | 0.85 | 0.811 | 0.989 | **0.93** | 0.753 | 0.764 | 0.82 |
| 5th iteration | 0.845 | 0.805 | 0.989 | 0.842 | 0.744 | 0.804 | 0.825 |
| **Average** | **0.861** | **0.826** | **0.989** | **0.878** | **0.771** | **0.805** | **0.875** |

(The best results of each parameter are marked in bold text.)

Table 8. The OA, Kappa score, and F1-score of KPConv classification (scenario 3)

|  | OA | κ | F1-score | | | | |
|---|---|---|---|---|---|---|---|
|  |  |  | Level 1 | Level 2 | Level 3 | Level 4 | Level 5 |
| 1st iteration | **0.848** | **0.809** | 0.854 | 0.8 | 0.817 | **0.851** | 0.925 |
| 2nd iteration | 0.767 | 0.704 | 0.738 | 0.533 | 0.723 | 0.822 | **0.952** |
| 3rd iteration | 0.805 | 0.755 | 0.844 | 0.767 | 0.762 | 0.767 | 0.931 |
| 4th iteration | 0.816 | 0.768 | 0.86 | 0.789 | 0.744 | 0.825 | 0.848 |
| 5th iteration | 0.846 | 0.806 | **0.905** | **0.875** | **0.835** | 0.784 | 0.841 |
| **Average** | **0.816** | **0.768** | **0.840** | **0.753** | **0.776** | **0.81** | **0.899** |

(The best results of each parameter are marked in bold text.)

Table 9. The OA, Kappa score, and F1-score of KPConv classification (scenario 4)

|  | OA | κ | F1-score | | | | |
|---|---|---|---|---|---|---|---|
|  |  |  | Level 1 | Level 2 | Level 3 | Level 4 | Level 5 |
| 1st iteration | **0.904** | **0.879** | 1 | **0.97** | **0.879** | 0.813 | 0.87 |
| 2nd iteration | 0.878 | 0.847 | 1 | 0.917 | 0.854 | 0.791 | 0.817 |
| 3rd iteration | 0.888 | 0.859 | 1 | 0.882 | 0.824 | 0.831 | 0.918 |
| 4th iteration | 0.893 | 0.865 | 0.989 | 0.935 | 0.822 | 0.824 | 0.920 |
| 5th iteration | 0.878 | 0.846 | 0.979 | 0.759 | 0.8 | **0.867** | **0.984** |

| | | | | | | |
|---|---|---|---|---|---|---|
| **Average** | **0.888** | **0.859** | **0.994** | **0.893** | **0.836** | **0.825** | **0.902** |

(The best results of each parameter are marked in bold text.)

Figure 15 shows the tree decay level classification result of a sample plot. The result is obtained by the best performed model trained in scenario 4.

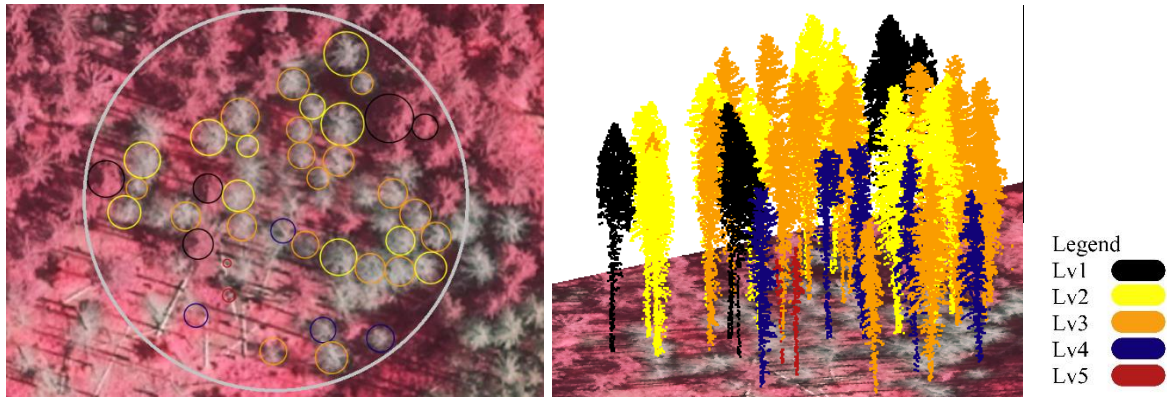

**Figure 15.** Classification result of the sample plot (left: top view; right: 3D view).

### 4.3.2 CNN

Table 10 summarizes the OAs, kappa scores, and F1-scores obtained by the CNN approach. In general, this approach attained an average OA reaching 88.4% and an average kappa score of 0.854 throughout our experiments. Among the five classes of tree decay levels, Decay Level 1 was classified as the best, reaching an average F1-score of 0.989, while Level 4 performed worst, reaching an average F1-score of 0.817.

**Table 10.** The OA, Kappa score, and F1-score of the CNN classification

| | OA | κ | F1-score | | | | |
|---|---|---|---|---|---|---|---|
| | | | **Level 1** | **Level 2** | **Level 3** | **Level 4** | **Level 5** |
| 1st iteration | **0.893** | 0.864 | 1 | 0.889 | 0.819 | 0.835 | **0.947** |
| 2nd iteration | 0.898 | **0.87** | 0.956 | 0.952 | **0.854** | 0.838 | 0.929 |
| 3rd iteration | 0.883 | 0.852 | 1 | 0.844 | 0.8 | **0.851** | 0.931 |
| 4th iteration | 0.854 | 0.816 | 1 | 0.941 | 0.785 | 0.75 | 0.8 |
| 5th iteration | **0.893** | 0.866 | 0.989 | **0.957** | 0.847 | 0.809 | 0.873 |
| **Average** | **0.884** | **0.854** | **0.989** | **0.917** | **0.821** | **0.817** | **0.896** |

(The best results of each parameter are marked in bold text.)

### 4.3.3 Random Forest

The RF approach achieved an average OA of 85.9% and an average kappa score of 0.822. Decay Level 1 attained the highest average F1-score reaching the perfection of 1, while Level 4 attained the lowest with (F1-score = 0.786).

Table 10. The OA, Kappa score, and F1-score of RF classification

| | OA | κ | F-1 score | | | | |
|---|---|---|---|---|---|---|---|
| | | | Level 1 | Level 2 | Level 3 | Level 4 | Level 5 |
| 1st iteration | 0.863 | 0.826 | 1 | 0.792 | 0.807 | 0.817 | 0.881 |
| 2nd iteration | **0.887** | **0.858** | 1 | **0.952** | 0.816 | 0.773 | 0.923 |
| 3rd iteration | 0.834 | 0.791 | 1 | 0.923 | **0.824** | 0.717 | 0.7 |
| 4th iteration | 0.844 | 0.803 | 1 | 0.781 | 0.727 | 0.786 | **0.966** |
| 5th iteration | 0.868 | 0.833 | 1 | 0.833 | 0.774 | **0.835** | 0.9 |
| **Average** | **0.859** | **0.822** | **1** | **0.856** | **0.79** | **0.786** | **0.874** |

(The best results of each parameter are marked in bold text.)

## 5. Discussion

In this study, we evaluated 2D- and 3D-based approaches for classifying five levels of Norway spruce decay stage using ML/DL. Our method obtained a very good accuracy for classifying different decay stages of Norway spruce trees. Overall, KPConv approach (scenario 4) outperformed (OA = 88.8%, κ = 0.856) in our experiments, followed by CNN (OA = 88.4%, κ = 0.853). Compared to these two, RF experiment had slightly lower performance (OA = 85.9%, κ = 0.822).

Based on the results generated by the four scenarios of KPConv classification, we can infer that using multispectral point clouds (X, Y, Z, I, NIR, R, G) as input led to better results than simply using X, Y, and Z values. This proved that the fusion of ALS point clouds and CIR images was crucial in classifying tree decay stages. From the results of four scenarios, there is a considerate increase on the classification accuracy of decay level 1 and 2 with color information. This showed the significance of color differences in CIR images which helped distinguish healthy trees and unhealthy trees even though they appeared similar in terms of their geometry and size. Scenario 3 implied the efficiency of intensity information compared to simply using coordinate values. In scenario 4, when considering all features, best results are achieved among the KPConv experiments.

In addition, the results showed that 2D- and 3D-based approaches had similar performance, which proves that both solutions are feasible for this task. The classification results based on point clouds are still slightly higher than based on images. A possible explanation is that the 3D-based approach maintains more precise and complete geometric features. Additionally, due to the distinctive color characteristics of trees of decay level 1, from Table 11, almost no errors are observed in this category for all methods.

**Table 11.** F1-score of RF, CNN and KPConv classification

| | OA | κ | F-1 score | | | | |
|---|---|---|---|---|---|---|---|
| | | | Level 1 | Level 2 | Level 3 | Level 4 | Level 5 |
| RF | 0.859 | 0.822 | 1 | 0.856 | 0.79 | 0.786 | 0.874 |
| CNN | 0.884 | 0.854 | 0.989 | **0.917** | 0.821 | 0.817 | 0.896 |
| KPConv (scenario 4) | **0.888** | **0.859** | 0.994 | 0.893 | **0.836** | **0.825** | **0.902** |

(The best results of each parameter are marked in bold text.)

Comparing the two image-based approaches, both achieved a promising classification accuracy on the image dataset. The higher overall accuracy attained by the CNN approach reflects the advantages of using DL for image classification over traditional ML. Apart from the difference in model architectures, the CNN model extracted features by itself, thus, the most distinctive features between different classes could be extracted automatically. While we only extracted three handcrafted features from the images for RF classification. In this case, the accuracy might still can be improved if more representative features were extracted. Overall, the image-based approaches have performed well in distinguishing the five decay stages of individual spruce trees.

This investigation is very significant in the sense that it could equally be viewed as a pilot study aiming toward automated approaches for large-scale (regional, national, and global levels) cost-effective forest monitoring and management, which would bring relief and sustainability comparing with the traditional field surveying approach. Over the past few decades, forest managers have been using remote sensing sensors for monitoring, inventory, and management, with advancements in remote sensing technology. Processing remote sensing data for large-scale forestry, especially for monitoring, would be very herculean without a sustainable robust automated approach because there will be the need for frequent and consistent processing of voluminous data, which DL can do within a reasonable time. However, traditional ML algorithms like the RF have a limit to the amount of data they can process, thus, it will not be cost-effective. Our results, for instance, show that both the 2D- and 3D-based DL approaches could be applied on forest monitoring and management. Notwithstanding, RF can be considered in cases of forest monitoring at plot level, where small amounts of data are to be processed.

## 6. Conclusion

Monitoring forest health is very critical for forest ecosystem management and environmental protection. Especially dead wood is an important indicator for assessing forest health and conversation as forest biodiversity is closely linked to the amount and quality of dead wood. Tree decay detection and classification are also very important for safety assessments of forests (Mattheck & Bethge, 1993) and analysis of nest webs of cavity-nesting species (Altamirano et al., 2017). In recent years, applications of remote sensing-based ML techniques in forest health assessment are rapidly increasing but they are still in their infancy. In this study, we developed a classification scheme of different tree decay stages from ALS point clouds and CIR images. Our results demonstrate the robustness of ML/DL algorithms in classifying single-tree decay stages of Norway spruce from airborne remote sensing platforms for the first time. We expect that our approach can also be easily transferred to other coniferous tree species as these classes of trees show similar properties and decay processes. Despite, the performance of KPConv being the best, it cannot be absolutely compared with the CNN or RF as they have different classification objects (point clouds and imagery respectively). Rather, the results for both KPConv and CNN further show the prowess of DL. Our experiments also revealed that the combination of ALS point cloud and CIR imagery had a significant impact on improving classification accuracy. Applying KPConv as well as CNN-based frameworks can open new opportunities for monitoring dead wood quality on a landscape scale, which is a crucial indicator of forest biodiversity. This will contribute to better evaluation of forest health, enabling more efficient and sustainable forest ecosystem management and conservation. Moreover, the method can be also applied to other tasks such as the assessment of trees close to roads and houses in regard of public safety.

**Disclosure statement**

No potential conflict of interest was reported by the author(s).

**Data availability statement**

The data that support the findings of this study are available from the corresponding author, [J-H. Wang, W.Y], upon reasonable request.

**Acknowledgement**

This work was supported by the National Natural Science Foundation of China (Project No. 42171361), the Research Grants Council of the Hong Kong Special Administrative Region, China, under Project PolyU 25211819. This work was partially supported by The Hong Kong Polytechnic University under Project 1-ZVN6, 1-ZECE.